\documentclass[11pt]{article}

\usepackage[preprint]{acl}

\usepackage{times}
\usepackage{latexsym}

\usepackage[T1]{fontenc}

\usepackage[utf8]{inputenc}

\usepackage{microtype}

\usepackage{inconsolata}

\usepackage{graphicx}

\newcommand\logo{\raisebox{-7pt}{\includegraphics[width=2em]{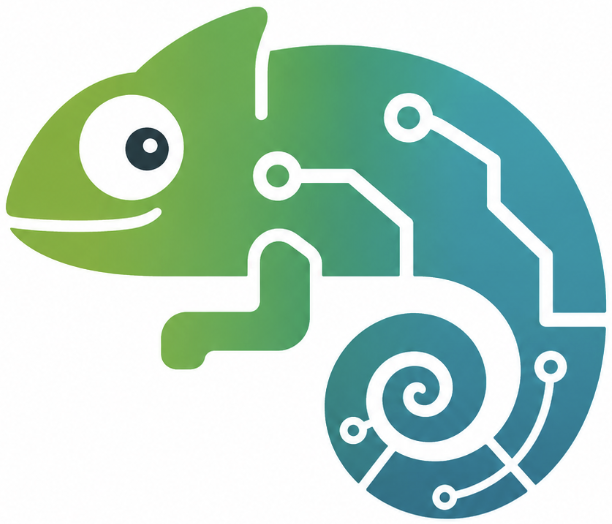}}}
\usepackage{colortbl}
\usepackage{enumitem}
\usepackage{multirow}
\usepackage{booktabs}
\usepackage{subfig}
\usepackage{pifont}
\usepackage{amsfonts, bbm}
\usepackage[most]{tcolorbox}

\definecolor{humanblue}{HTML}{48CCFC}
\definecolor{clusterA}{HTML}{FDC131} 
\definecolor{clusterB}{HTML}{558E28} 
\definecolor{clusterC}{HTML}{A072FD} 
\definecolor{clusterD}{HTML}{E965A4} 

\usepackage{xspace}
\newcommand{\circone}{\ding{172}\xspace}
\newcommand{\circtwo}{\ding{173}\xspace}
\newcommand{\circthree}{\ding{174}\xspace}

\usepackage{cleveref}
\crefname{figure}{Fig.}{Figs.}
\crefname{section}{Sec.}{Secs.}
\crefname{table}{Tab.}{Tabs.}

\title{{\logo}The Chameleon’s Limit: Investigating Persona Collapse and Homogenization in Large Language Models}

\author{
Yunze Xiao$^{1\dagger}$ \ \
Vivienne J. Zhang$^{2\dagger}$ \ \
Chenghao Yang$^{2}$ \ \
Weihao Xuan$^{3,4}$ \ \
Jen-tse Huang$^{5\ddagger}$
\\
$^{1}$CMU \quad 
$^{2}$UChicago \quad
$^{3}$UTokyo \quad
$^{4}$RIKEN AIP \quad
$^{5}$JHU \\
{\small $^{\dagger}$Equal contribution. \quad $^{\ddagger}$Corresponding Author: \texttt{jhuan236@jhu.edu}} \\
\parbox{0.03\textwidth}{\includegraphics[width=\linewidth]{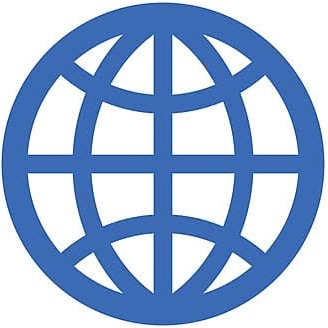}}\ \href{https://algoroxyolo.github.io/projects/chameleon-limit/}{\small\texttt{Website}}
\qquad
\parbox{0.03\textwidth}{\includegraphics[width=\linewidth]{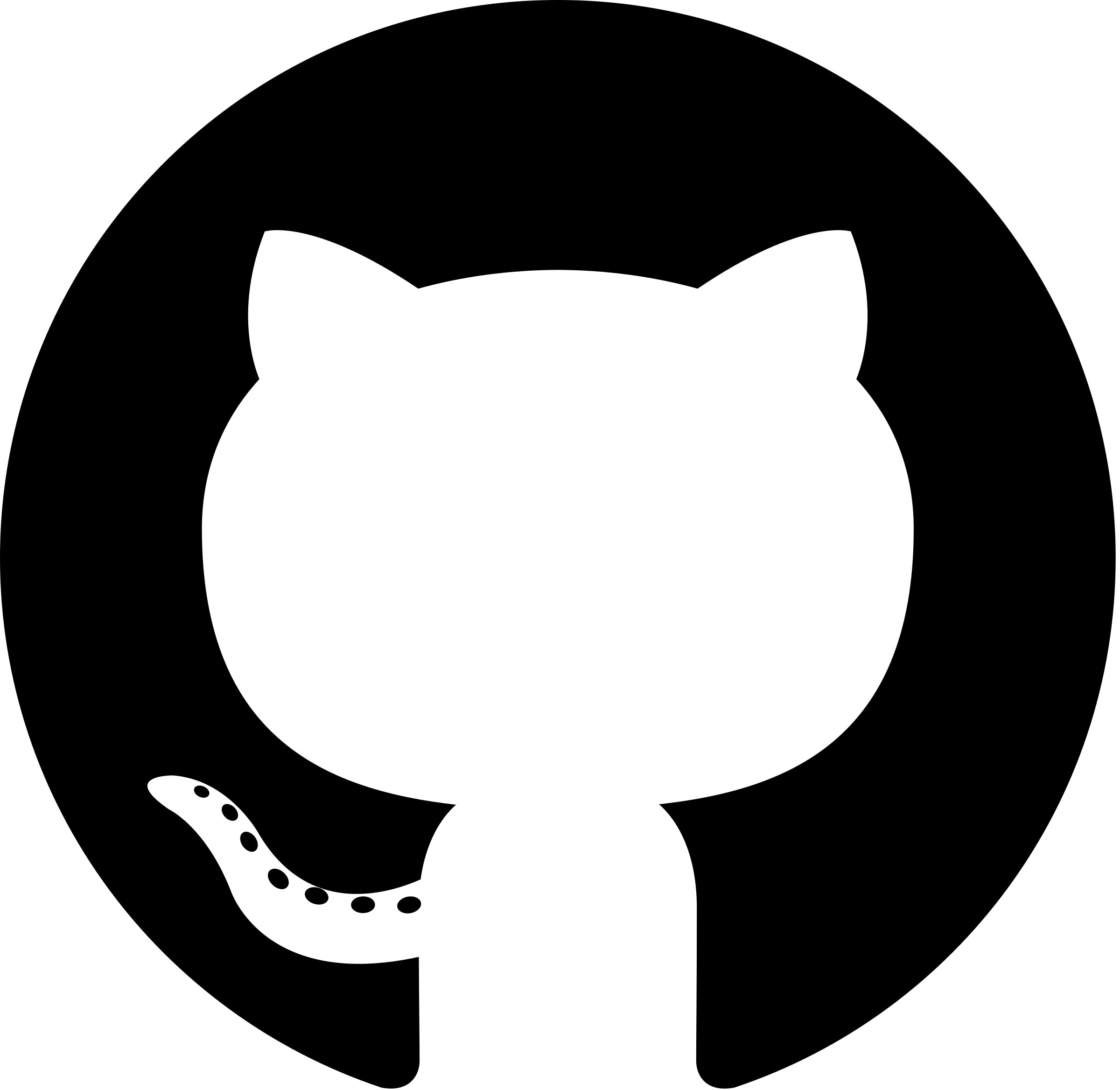}}\ \href{https://github.com/Algoroxyolo/PersonaCollapse}{\small\texttt{Codebase}}
}

\begin{document}
\maketitle
\begin{abstract}
Large language models (LLMs) have powered multi-agent simulations such as survey respondents or application users.
However, LLMs lack the population-level diversity required by these simulations---responses from LLMs with detailed human profiles cluster into classes, unlike human responses that uniformly fill the entire space.
In this paper, we formalize this as \textit{Persona Collapse} and quantify its population-level structure with three diagnostic metrics: Coverage (how much of the persona space is occupied), Uniformity (how evenly agents spread across it), and Complexity (how rich the resulting behaviors are).
Across eight LLMs on personality (BFI-44), moral reasoning, and self-introduction, persona collapse is universal but task-dependent: CoSER-Llama-8B collapses the most in personality while being the most diverse in moral reasoning.
Moreover, sampling more personas cannot mitigate the collapse.
We further show a fidelity--diversity tradeoff: the models with the highest per-persona fidelity produce the most polarized populations.
Finally, persona-targeted fine-tuning further narrows the population instead of diversifying it.
\end{abstract}

\section{Introduction}

\begin{figure}[t]
    \centering
    \includegraphics[width=1.0\linewidth]{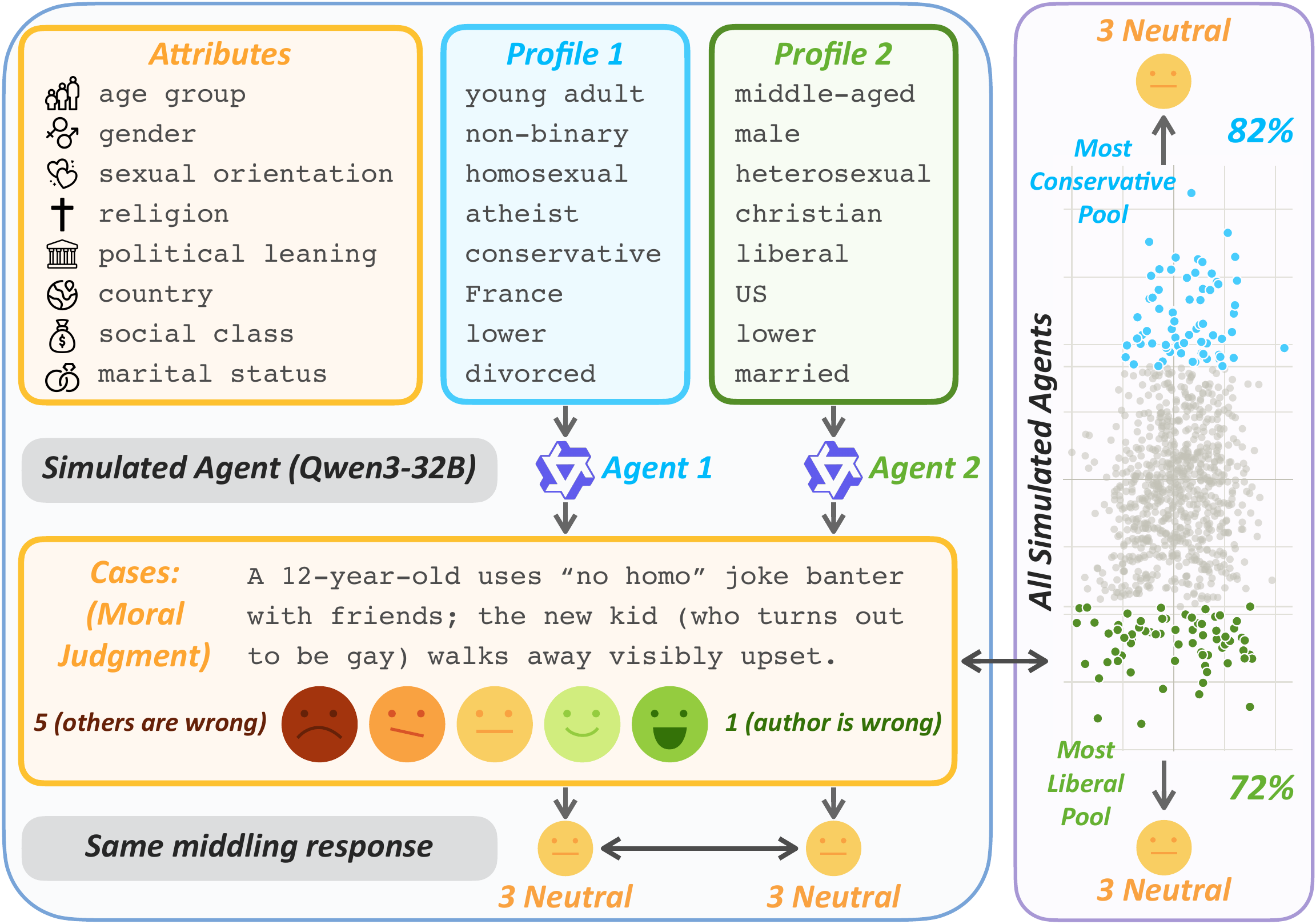}
    \caption{
    \textit{Persona Collapse} in LLM-based population simulation.
    Although two personas differ across dimensions, two Qwen3-32B agents with the profiles assign the same response (3: neutral) on a socially sensitive judgment task.
    At the population level, the most conservative and most liberal persona pools also concentrate on the same middle rating.
    }
    \label{fig:teaser_new}
\end{figure}

\begin{figure}[t]
  \includegraphics[width=1.0\linewidth]{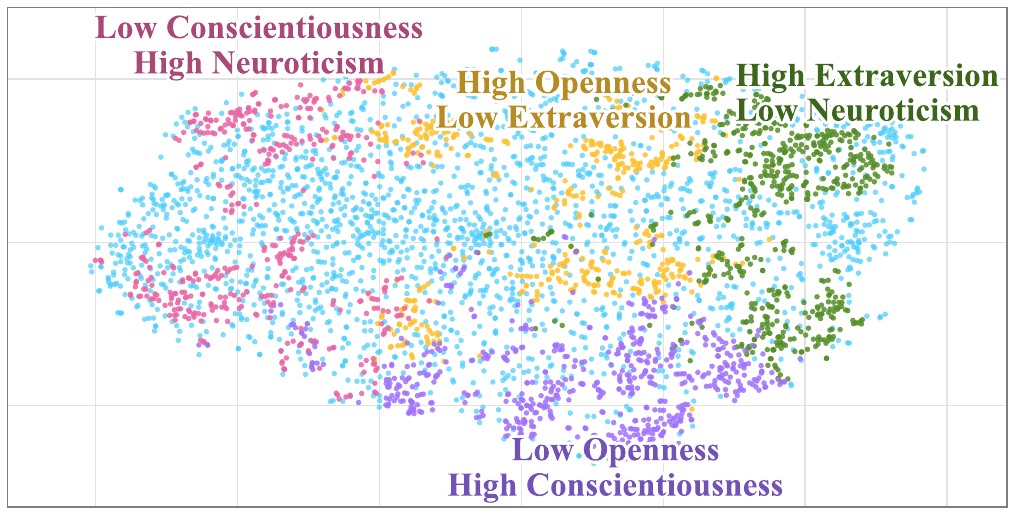}
  \caption{t-SNE projection of the BFI-44 responses using the same 2,058 profiles from Twin-2K-500~\citep{toubia2025database}. \textbf{\textcolor{humanblue}{Blue}}: Humans spread evenly across the space. \textbf{\textcolor{clusterA}{Yellow}, \textcolor{clusterB}{Green}, \textcolor{clusterC}{Purple}, and \textcolor{clusterD}{Pink}}: When assigned personas, Qwen3-32B responses fragment into separated clusters rather than filling the space.}
  \label{fig:teaser}
\end{figure}

Agents powered by large language models (LLMs) are increasingly deployed as participants in simulated societies, synthetic survey respondents, and proxies for user testing~\citep{park2023generative, zhou2025pimmur, louie2024roleplay}.
These applications rely on a critical assumption: given a richly specified persona (e.g., age, gender, nationality, political leaning, occupation), the model can simulate an individual whose behavior reflects the complex intersection of all these attributes~\citep{xiao2025humanizing}.
However, existing evaluations~\citep{wang2024incharacter, abdulhai2025consistently} primarily measure ``\textit{shallow fidelity}'', which assesses whether an agent successfully mimics a simple, low-dimensional label, such as a well-known public figure (e.g., ``Albert Einstein'') or a broad demographic prototype (e.g., ``a sixty-year-old Black woman'').
While such shallow fidelity works for role-playing, it remains unclear whether LLMs actually maintain fidelity for more complicated simulation targets or whether they might overlook certain information.

In this study, we identify a fundamental failure in such ``\textit{combinatorial fidelity}.''
When instructed to role-play a persona defined by $26$ dimensions, LLMs systematically retain only the most stereotypically salient attributes for downstream tasks, discarding the rest.
This attribute truncation propagates from individual personas to the population as a whole.
As illustrated in \cref{fig:teaser_new}, personas differing across multiple dimensions can nevertheless produce the same model judgment, causing the population that should diverge to concentrate on similar responses.
At the population level, model-generated data points occupy restricted and clustered regions rather than evenly spanning the human distribution in the 44-dimensional BFI personality space (\cref{fig:teaser}).
We call this population-level homogenization \textit{Persona Collapse}.

To systematically measure persona collapse, we shift from individual-level fidelity metrics to population-level diagnostics.
We formalize our analysis through the \textit{Behavioral Trait Matrix} $\mathbf{B} \in \mathbb{R}^{N \times D}$, where each row holds a single persona's responses to the $D$ behavioral items.
A structurally healthy simulated population should:
(i) span the full distribution of human data points rather than over-sampling a small region (\textbf{Coverage};~\citealp{naeem2020reliable});
(ii) distribute evenly across the space instead of collapsing into a few dense, degenerate clusters (\textbf{Uniformity};~\citealp{hopkins1954new, wang2020understanding}); and
(iii) be genuinely high-dimensional rather than compressed onto a low-dimensional subspace (\textbf{Complexity};~\citealp{levina2004maximum, tulchinskii2023intrinsic}).
Together, these three criteria form a diagnostic framework grounded in the geometry of $\mathbf{B}$.
Coverage measures how much of the behavioral space is occupied, Uniformity how evenly the population occupies it, and Complexity how many independent dimensions underlie that variation.
Since these population-level metrics do not reveal where collapse lives, we complement them by localizing which behavioral items are flattened and which persona attributes still drive variation~\citep{tucker1951method, john1999big}.

We apply this framework to eight LLMs across three tasks: personality simulation (BFI-44~\citep{john1999big}), moral reasoning~\citep{liu2025synthetic}, and free-form self-introductions.
Model responses to the BFI-44 are compared to a human reference from \citet{toubia2025database}.
Two findings characterize this behavior:
(i) models with the highest per-persona fidelity consistently produce the most polarized populations; and
(ii) a model can be structurally degenerate in personality simulation while remaining comparatively diverse in moral reasoning.
Our contributions are as follows:
\begin{enumerate}[nosep]
    \item[\circone] We propose a geometric diagnostic framework to systematically measure persona collapse, quantifying the loss of high-dimensional human attributes in LLM simulations.
    \item[\circtwo] We evaluate eight LLMs across personality simulation, moral reasoning, and open-ended self-introduction, revealing pervasive persona collapse across all models.
    \item[\circthree] We show a fidelity--diversity tradeoff and find that, in the two pipelines we study, persona-targeted fine-tuning does not mitigate persona collapse but instead exacerbates it.
\end{enumerate}
\section{Methods}
\label{sec:framework}

\begin{figure*}[t]
\centering
  \includegraphics[width=1.0\linewidth]{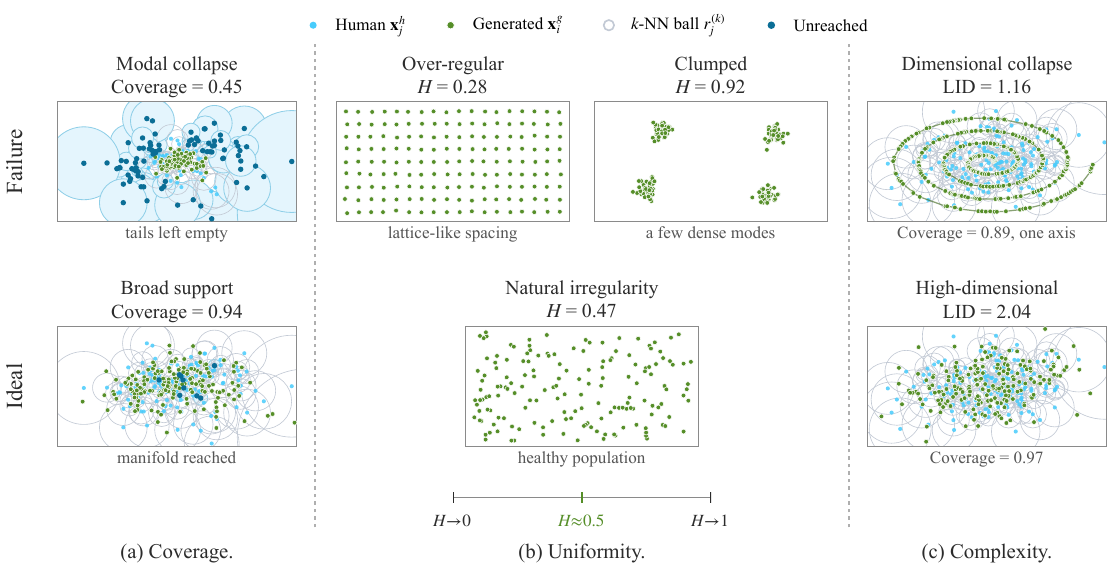}
  \caption{
  Illustrations of the three metrics.
  The first row shows the failures each metric can locate, while the second row shows the ideal cases.
  \textbf{(a)} A low coverage indicates that agents concentrate in a small region.
  \textbf{(b)} Natural distributions resemble spatial randomness ($H \approx 0.5$); a low uniformity indicates that results are over-engineered into lattices, while a high uniformity indicates that they degenerate into clusters.
  \textbf{(c)} A high coverage cannot entail a dimension-rich distribution; a low complexity indicates that agents collapse onto a lower-dimensional manifold.
  }
  \label{fig:three_axes_conceptual}
\end{figure*}

\subsection{Behavioral Trait Matrix}

We represent a population of $N$ personas as a Behavioral Trait Matrix $\mathbf{B}\in\mathbb{R}^{N \times D}$, whose rows are personas' responses to the $D$ items (questions) of an instrument; items aggregate into psychologically meaningful factors (e.g., the five BFI traits).
Let $X_g=\{\mathbf{x}^{g}_i\}_{i=1}^{N_g}$ and $X_h=\{\mathbf{x}^{h}_j\}_{j=1}^{N_h}$ denote the generated personas and the human reference data points, respectively, embedded in a shared space with Euclidean distance $\|\cdot\|_2$.
The rows of $\mathbf{B}$ form the point cloud $X_g$.
For a point $\mathbf{x}$ and a finite point set $A$, let $\operatorname{dist}(\mathbf{x},A)=\min_{\mathbf{y}\in A\setminus\{\mathbf{x}\}}\|\mathbf{x}-\mathbf{y}\|_2$ be the distance from $\mathbf{x}$ to $A$, and let $r_\ell(\mathbf{x};A)$ be the distance from $\mathbf{x}$ to its $\ell$-th nearest neighbor in $A\setminus\{\mathbf{x}\}$; hence $r_1(\mathbf{x};A)=\operatorname{dist}(\mathbf{x},A)$.

\subsection{Quantifying Persona Collapse}
\label{sec:axes}

\paragraph{Coverage} asks whether the population spans the behavioral space.
Following \citet{naeem2020reliable}, we build a hypersphere around each human reference point $\mathbf{x}^{h}_j$ with radius $r_k(\mathbf{x}^{h}_j;X_h)$.
Coverage is the fraction of these human neighborhoods that contain at least one generated persona:
\begin{equation}
\small
\operatorname{Cov}=\frac{1}{N_h}\sum_{j=1}^{N_h}\mathbbm{1}\!\bigl[
\operatorname{dist}(\mathbf{x}^{h}_j,X_g)\le r_k(\mathbf{x}^{h}_j;X_h)
\bigr].
\label{eq:coverage}
\end{equation}
We anchor to a human reference rather than the theoretical combinatorial space because genuine human variation occupies a restricted manifold, and exhaustively mapping the full space is intractable.
Low Coverage indicates that the model over-samples a small region (\cref{fig:three_axes_conceptual}a).

\paragraph{Uniformity} asks whether agents spread \textit{evenly} across the region they occupy.
Broad Coverage does not rule out a population that separates into a few dense clusters.
We quantify this using the Hopkins statistic~\citep{hopkins1954new}.
We sample $m$ generated points $\mathbf{x}^{g}_{s_1},\dots,\mathbf{x}^{g}_{s_m}$ from $X_g$ and $m$ random probes $\mathbf{z}_1,\dots,\mathbf{z}_m$ uniformly from the bounding region of $X_g$:
\begin{equation}
\small
H = \frac{\sum_{t=1}^{m} \operatorname{dist}(\mathbf{z}_t,X_g)}{\sum_{t=1}^{m} \operatorname{dist}(\mathbf{z}_t,X_g) + \sum_{t=1}^{m} \operatorname{dist}(\mathbf{x}^{g}_{s_t},X_g)}.
\label{eq:hopkins}
\end{equation}
If personas are distributed with natural irregularity, $H \approx 0.5$.
Deviations reveal distinct pathologies: clustering drives $H \to 1$,
whereas over-regular, lattice-like spacing drives $H \to 0$ (\cref{fig:three_axes_conceptual}b).

\paragraph{Complexity} asks whether the variation is high-dimensional.
There is a blind spot in Coverage: a low-dimensional manifold can connect enough neighborhoods to score well.
As shown in \cref{fig:three_axes_conceptual}c, a single line threads through most human neighborhoods, yielding a Coverage of $0.89$.
Complexity closes this gap by estimating the Local Intrinsic Dimensionality (LID) at each point $\mathbf{x}\in X$, $X\in\{X_g,X_h\}$, with the maximum likelihood estimator of \citet{levina2004maximum} over its $k$ nearest neighbors:
\begin{equation}
\small
\widehat{\operatorname{LID}}_k(\mathbf{x};X) = - \left( \frac{1}{k-1} \sum_{\ell=1}^{k-1} \log \frac{r_\ell(\mathbf{x};X)}{r_k(\mathbf{x};X)} \right)^{-1}.
\label{eq:lid}
\end{equation}
LID measures how fast the local neighborhood volume grows with radius: high values indicate genuinely high-dimensional local geometry, whereas $\widehat{\operatorname{LID}}_k \to 1$ indicates a neighborhood that is essentially linear.
\citet{tulchinskii2023intrinsic} also found that human-authored text carries systematically higher intrinsic dimensionality than LLM-generated text, motivating our design.

\subsection{Localizing Collapse}
\label{sec:item-diagnostics}

The three axes describe the overall shape of the generated population: how much of the human space it covers, how evenly, and in how many dimensions.
They do not tell us where the collapse happens: which items every persona answers the same way, whether the remaining differences are just group stereotypes, and which assigned attributes still affect behavior.
To answer these questions, we examine the response matrix $\mathbf{B}$ item by item, together with the attributes assigned to each persona.
Throughout, $q=1,\dots,Q$ indexes items, each answered on an $L$-point Likert scale, and the subscripts $g$ and $h$ denote the generated and human populations, consistent with \cref{sec:axes}.

\paragraph{Response range.}
For each item $q$, let $p_{q,c}$ be the fraction of personas choosing option $c\in\{1,\dots,L\}$.
We measure how many options the population effectively uses with $E_q = 1/\sum_{c=1}^{L} p_{q,c}^2$ (the inverse Simpson index).
$E_q$ ranges from 1, when all personas give the same answer, to $L$, when all options are used equally often.
A value near 1 means the model has erased individual differences on that item.

\paragraph{Demographic clustering.}
Next, we ask whether the remaining differences between personas reflect individuals or group stereotypes.
For each item, we compute $R^2$, the share of response variance explained by a single demographic attribute (e.g., gender or political ideology).
In human data, $R^2$ is typically small, because people differ much more within groups than between them~\citep{costa1992revised}.
A higher $R^2$ in the generated population means the model treats personas as members of their group rather than as individuals.
Averaging $R^2$ over items shows which attributes cause the most distortion, and in which traits.

\paragraph{Attribute dominance.}
Finally, we ask which assigned attributes the model actually uses.
We add the four attributes to a regression one at a time (political ideology $\to$ gender $\to$ country $\to$ social class) and record how much each one increases $R^2$.
We summarize the result with Dom\%, the share of the total $R^2$ contributed by the strongest attribute; if all four attributes mattered equally, Dom\% would be 25\%.
For open-ended self-introductions, we report two further measures.
The mention rate (MR) is the fraction of self-introductions that explicitly state each assigned attribute, detected by keyword matching.
The intra-class correlation (ICC) measures how much of the variation in linguistic features comes from differences between personas rather than from differences among a persona's three samples.
MR is averaged over gender, country, political ideology, age, and social class, and ICC over seven linguistic features.

\paragraph{Persona fidelity.}
We check whether personas actually express their assigned trait levels.
For each trait, a persona's trait score is the mean of its responses to that trait's items.
Fidelity $\rho$ is Spearman's rank correlation between the assigned level (Low/Neutral/High, coded 1/2/3) and the trait score across all $N_g$ personas.
As a complementary effect size, we compute Cohen's $d$ between High and Low personas on each trait and report its average over the five traits, $\bar d$.
Both measures require an assigned target level, so they are undefined for the human reference, whose participants were never assigned one.
Instead, we compare against typical effect sizes in the personality literature: observed Big Five group differences are usually $|d|\approx0.1$--$0.5$, whereas $d\approx1.2$ already counts as ``very large'' and $d\approx2$ as ``huge''~\citep{costa1992revised, sawilowsky2009new}.

\subsection{Experimental Settings}
\label{sec:setup}

\paragraph{Datasets.}

We evaluate the simulated personas across three tasks:
First, the Big Five Inventory (BFI-44; \citealp{john1999big}) uses 44 5-point Likert items to measure five established personality traits, enabling tests of behavioral diversity.
Second, we use 131 ethical scenarios from a moral reasoning dataset \citep{liu2025synthetic}, also rated on a 5-point scale.
Finally, personas produce three open-ended self-introductions, which expose failure modes missed by forced-choice tasks, such as template homogenization, and test whether collapse carries over to generative settings.
The prompts are provided in \cref{app:prompts}.

\paragraph{Models.}

We evaluate eight instruction-tuned LLMs in two categories.
The general-purpose set includes Llama-3.1-8B-Instruct \citep{llama31}, Qwen3-4B, Qwen3-30B-A3B, and Qwen3-32B \citep{qwen3}; Qwen3 models run in their default thinking mode, and we additionally report Qwen3-32B with thinking disabled.
The role-play set includes CoSER-Llama-8B \citep{wang2025coser}, fine-tuned from Llama-3.1-8B-Instruct on literary character data; HER-32B \citep{du2026her}, trained from Qwen3-32B with persona-specific supervised fine-tuning (PSFT) followed by reinforcement learning; CoSER-Qwen-32B, which we train ourselves to approximate HER-32B's intermediate PSFT-only stage; and MiniMax-M2-Her \citep{minimax-m2her}, a large-scale role-play model released by MiniMax.
This selection yields two paired comparisons on a shared backbone:
(i) Llama-3.1-8B-Instruct $\to$ CoSER-Llama-8B isolates the effect of PSFT;
(ii) Qwen3-32B $\to$ CoSER-Qwen-32B $\to$ HER-32B approximates successive post-training stages (PSFT, then RL).
Since every starting point is already post-trained, these pairs measure the effect of additional persona-targeted training rather than of alignment itself.
All models are queried through the OpenRouter API with all decoding hyperparameters (temperature, top-$p$, etc.) left at their API defaults unless otherwise noted.

\begin{figure*}[t]
\centering
\includegraphics[width=1.0\linewidth]{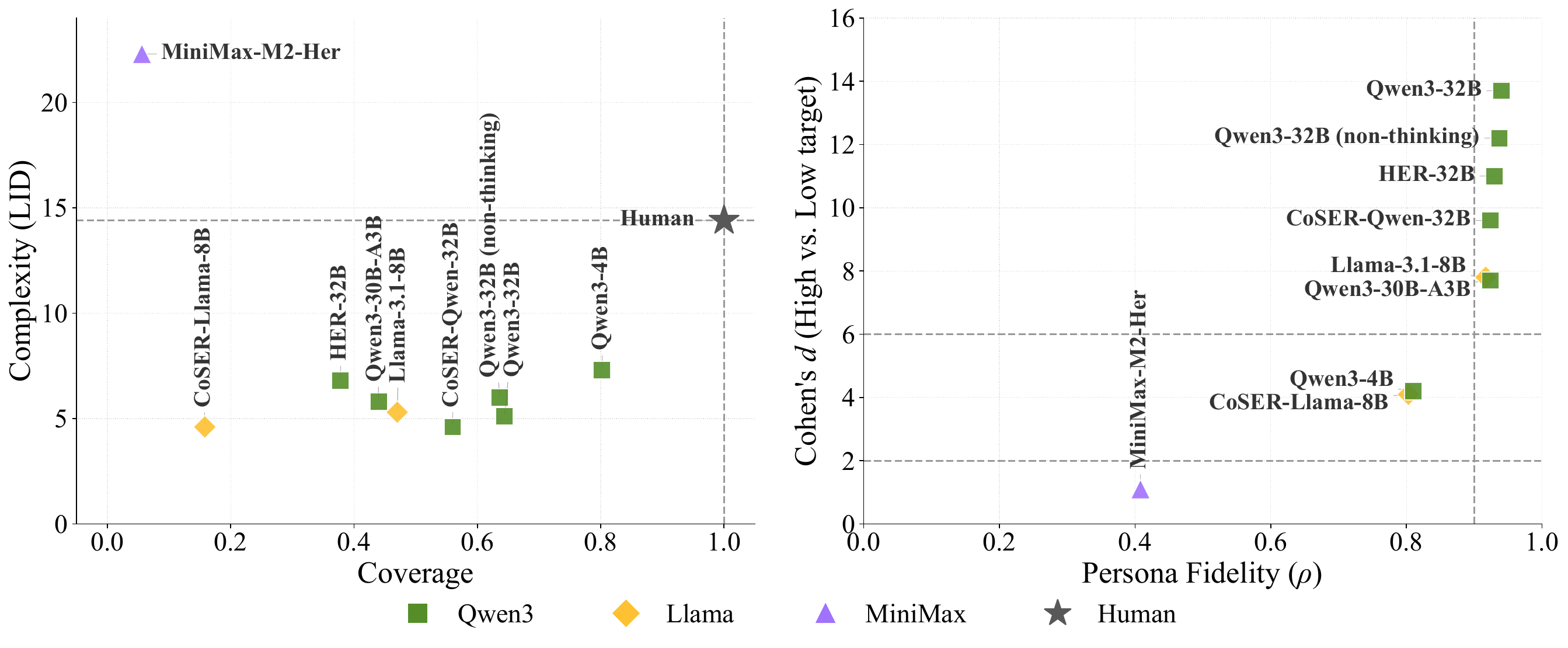}
\caption{Population-level diagnostics on BFI-44 (8 models, 1,144 personas each). \textbf{Left}: Cov vs. LID. No model simultaneously achieves high coverage and high complexity to reach the human reference performance. \textbf{Right}: Persona fidelity ($\rho$) vs. trait polarization (Cohen's $d$). Every model with $\rho > 0.9$ produces $d > 6$, far exceeding natural human group differences ($d\approx2$ is typically considered ``huge'' in human personality research). $\rho$ and $d$ apply only to the prompted model populations, not to the unprompted human reference (\cref{sec:item-diagnostics}).}
\label{fig:combined}
\end{figure*}

\paragraph{Persona profiles and human reference.}

We compile 26 persona dimensions frequently used in prior work \citep{zhou2024sotopia, huang2025visbias, samuel2025personagym}, spanning demographics (e.g., age, gender, country), psychographic traits (e.g., political ideology), and individualized characteristics (e.g., hobbies, physical appearance).
We construct personas bottom-up by sampling attribute combinations, instead of relying on any single survey or asking LLMs to characterize a target population directly.
This is a deliberate stress test: it presents the model with diverse, intersecting, and sometimes non-typical individuals and asks whether the simulated population preserves that diversity, rather than what the model is most likely already optimized to reproduce.
We initially sample 2,000 attribute combinations from this space; after manual screening for completeness and internal consistency, 856 are excluded, leaving 1,144 personas for our three tasks (see \cref{app:persona} for the full attribute list and filtering criteria).
We use human BFI-44 responses from Twin-2K-500 (2,058 participants; \citealp{toubia2025database}) as the reference distribution; the other two instruments focus on model-to-model comparisons and item-level diagnostics (\cref{sec:item-diagnostics}).
\section{Results}
\label{sec:results}

\begin{table*}[t]
\centering
\resizebox{0.96\textwidth}{!}{
\begin{tabular}{l rrrrrr rrr rrrr}
\toprule
& \multicolumn{6}{c}{\textbf{Personality (BFI-44)}} & \multicolumn{3}{c}{\textbf{Moral Reasoning}} & \multicolumn{4}{c}{\textbf{Self-Introduction}} \\
\cmidrule(lr){2-7} \cmidrule(lr){8-10} \cmidrule(lr){11-14}
\textbf{Model} & Cov & $H$ & LID & $\bar{E}$ & $\rho$ & $\bar{d}$ & LID & $\bar{E}$ & $\bar{R}^2_{\mathrm{pol}}$ & $H$ & LID & MR & ICC \\
\midrule
\textit{Human} & \textit{1.00} & \textit{0.57} & \textit{14.4} & \textit{3.69} & --- & --- & --- & --- & --- & --- & --- & --- & --- \\
\midrule
Llama-3.1-8B & 0.47 & 0.69 & 5.3 & 3.36 & .92 & 7.8 & 41.8 & 2.99 & .004 & 0.66 & 16.6 & .60 & .33 \\
Qwen3-4B & \textbf{0.80} & 0.71 & \textbf{7.3} & 2.98 & .81 & 4.2 & 15.1 & 1.20 & \textbf{.005} & 0.67 & 17.5 & .62 & \textbf{.51} \\
Qwen3-30B-A3B & 0.44 & 0.68 & 5.8 & 3.02 & .92 & 7.7 & 32.3 & 2.01 & .003 & 0.66 & 20.2 & .66 & .41 \\
Qwen3-32B & 0.64 & 0.70 & 5.1 & 3.23 & \textbf{.94} & \textbf{13.7} & 31.7 & 1.83 & .002 & 0.65 & 20.8 & \textbf{.69} & .34 \\
Qwen3-32B (non-thinking) & 0.64 & 0.67 & 6.0 & 3.32 & \textbf{.94} & 12.2 & 34.9 & 2.01 & .003 & --- & --- & --- & --- \\
\midrule
CoSER-Llama-8B & 0.16 & 0.91 & 4.6 & 1.36 & .80 & 4.1 & \textbf{45.3} & \textbf{4.27} & .003 & 0.67 & 21.4 & .53 & .28 \\
CoSER-Qwen-32B & 0.56 & 0.72 & 4.6 & 2.35 & .92 & 9.6 & 28.4 & 1.95 & .003 & 0.66 & 19.5 & .41 & .23 \\
HER-32B & 0.38 & 0.65 & 6.8 & \textbf{3.58} & .93 & 11.0 & 38.5 & 3.21 & .003 & 0.65 & 25.5 & .65 & .22 \\
MiniMax-M2-Her & 0.06 & \textbf{0.50} & 22.3 & 4.49 & .41 & 1.1 & 42.6 & 2.86 & .003 & 0.64 & \textbf{26.5} & .38 & .17 \\
\bottomrule
\end{tabular}
}
\caption{BFI-44 personality and moral reasoning ($N_g{=}1{,}144$ personas per model; human reference: $N_h{=}2{,}058$) results. Upper block: general-purpose models; lower block: role-play models. \textbf{Bold}: closest to human for Cov, $H$, LID, and $\bar{E}$ on BFI-44; highest value in all other columns. Self-introduction $H$ and LID are computed on persona embeddings: each sample is truncated to 512 characters and encoded with \texttt{all-MiniLM-L6-v2}~\citep{reimers2019sentence}, and the three samples of a persona are averaged into one vector.}
\label{tab:geometric}
\end{table*}

We organize our findings around four research questions (RQs):
\textbf{RQ1} (\textit{Population Geometry}): Do simulated populations evenly fill the human behavioral space?
\textbf{RQ2} (\textit{Localizing Collapse}): At which items and attributes does individual variation disappear?
\textbf{RQ3} (\textit{Fidelity vs. Diversity}): Does following personas more faithfully make the population more human-like?
\textbf{RQ4} (\textit{Post-Training}): Does persona-targeted training reduce collapse?
\cref{tab:geometric} reports core metrics defined in \cref{sec:framework}.

\subsection{RQ1: Population Geometry}
\label{sec:rq_geometry}

\paragraph{Most models cluster more than humans.}
On personality, every model except MiniMax-M2-Her has a higher Hopkins statistic than humans ($H$ from $0.65$ to $0.91$, vs. $0.57$).
The simulated population does not become a human-like cloud: it sits in a few piles near the human center, and the tails stay empty.
CoSER-Llama-8B is the most extreme case ($H{=}0.91$, Cov${=}0.16$).

\paragraph{High Coverage does not mean high Complexity.}
Coverage varies widely across models ($0.06$ to $0.80$), but even the models that reach many human neighborhoods stay far below the human LID ($14.4$).
Qwen3-4B is the clearest case: the best Coverage we see ($0.80$), yet LID is only half the human value ($7.3$).
Spreading personas across the map is not the same as reproducing its structure.

\paragraph{High Complexity can still lie outside the human range.}
This failure is easy to misread as a success: the population can be rich and even, yet lie almost entirely outside the human region.
MiniMax-M2-Her is that case (LID${=}22.3$, $H{=}0.50$, Cov${=}0.06$).
Without Coverage, it would look like the best system.

\begin{figure*}
    \centering
    \includegraphics[width=1.0\linewidth]{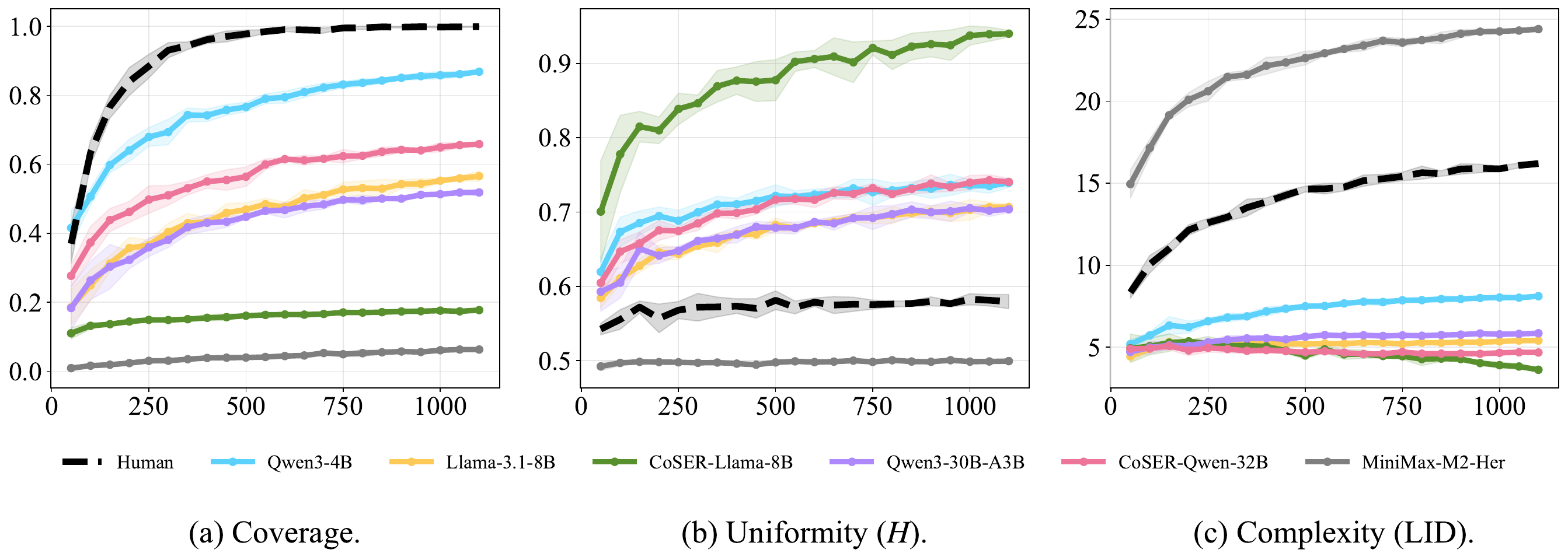}
    \caption{BFI-44 diagnostics vs. population size $N$. For each $N$ from 50 to 1,100 (step 50), we sample $N$ personas five times and plot the mean; shaded bands show the 95\% confidence interval across the five draws.}
    \label{fig:scale-up}
\end{figure*}

\paragraph{More personas do not close the gap.}
One might attribute these gaps to a finite sample: with enough personas, the population could eventually fill the human space.
\cref{fig:scale-up} tests this by subsampling $N$ from $50$ to $1{,}100$ personas.
Humans saturate quickly while models grow slowly and stay far below.
As $N$ grows, models gain little Coverage or LID and drift further from $H{=}0.5$.
Persona collapse is therefore a property of the model rather than of the sample size, and supplying more personas (e.g., as in PersonaHub-1B~\citep{ge2024scaling} and MatrAIx-8B\footnote{\url{https://matraix.ai/}}) does not fix it.

\paragraph{The same model's geometry changes with the task.}
A model that collapses on one task can be among the most complex on another.
CoSER-Llama-8B is the most collapsed model on personality (Cov${=}0.16$, $H{=}0.91$), yet it has the highest LID on moral reasoning ($45.3$) and the third highest of eight on self-introduction ($21.4$).
Qwen3-4B shows the reverse pattern: the highest BFI Coverage ($0.80$), but the lowest moral LID ($15.1$).

\subsection{RQ2: Localizing Collapse}
\label{sec:rq_localizing}

\paragraph{Models use fewer answer options than humans.}
On personality, every model except MiniMax-M2-Her uses a narrower answer scale than humans ($\bar{E}$ from $1.36$ to $3.58$, vs. $3.69$).
CoSER-Llama-8B is the most extreme case: it answers only $\{1, 3, 5\}$, $83.7\%$ of its answers are ``3,'' and $\bar{E}$ falls to $1.36$.
As with geometry, response range depends on the task.
The same CoSER-Llama-8B has the widest response range of any model on moral reasoning ($\bar{E}{=}4.27$), while Qwen3-4B moves the other way, from $\bar{E}{=}2.98$ on personality to the narrowest moral range ($\bar{E}{=}1.20$, with $17/131$ zero-variance items).

\paragraph{Demographic attributes explain little of the moral variance.}
On moral items, political ideology explains at most $0.5\%$ of the response variance in any model ($\bar{R}^2_{\mathrm{pol}}$ from $0.002$ to $0.005$).
The variation that remains is therefore barely tied to the assigned political identity: on moral reasoning, collapse does not take the form of sorting personas into political camps.

\paragraph{No single attribute dominates, and the one that matters changes with the task.}
The small demographic variance that does exist is usually spread across attributes rather than concentrated on one.
Measured by Dom\%, the share of demographic $R^2$ contributed by the strongest attribute, only Qwen3-4B leans hard on a single axis, Social Class ($59\%$); the remaining seven models sit between $29\%$ and $41\%$, against an even split of $25\%$.
Which attribute a model relies on also depends on the task: Qwen3-4B shifts from class on moral items ($59\%$) to gender in self-introductions ($61\%$; \cref{tab:text_r2} in \cref{app:selfintro-analysis}).
Where a dominant axis appears, it may reflect training data or alignment procedures, but since training details are not fully transparent, we refrain from causal attribution.
In extreme cases, stereotypical priors manifest not as statistical bias but as outright refusal (\cref{sec:appendix_failures}).

\paragraph{Models mention gender and country, but drop age and class.}
When a model compresses a 26-attribute persona into free text, it does not retain attributes evenly: easy, visible labels survive, while age and class drop out.
A simulated society built this way will look mixed on gender and nationality, and empty on socioeconomic life.
Averaged over the eight models, the mention rates follow a stable order: Gender ($89\%$) $>$ Country ($86\%$) $>$ Politics ($57\%$) $>$ Age ($30\%$) $>$ Social Class ($23\%$).
No model names Social Class in more than $42\%$ of introductions.
Details of the keyword detection are in \cref{app:layer1}, and per-model rates are in \cref{sec:appendix_mention}.

\paragraph{In free text, a persona is either stable but templated, or not stable at all.}
Free text exposes two opposite failures that Likert scores cannot.
Qwen3-4B is the most self-consistent model (ICC${=}0.51$), but its consistency comes with rigid, templated introductions (\cref{sec:appendix_template}).
MiniMax-M2-Her is the opposite: its self-introductions are the most complex we measure (LID${=}26.5$), but ICC is only $0.17$, so $83\%$ of its linguistic variation is stochastic noise.
Three samples of the same persona differ almost as much as samples from different personas.
Variety without a self is not a population.
Taken together, the failures in RQ1 and RQ2 differ across survey and free-text formats, so a persona evaluation that uses only one format can miss the failure the other format shows.

\subsection{RQ3: Fidelity vs. Diversity}
\label{sec:rq_fidelity}

\paragraph{Higher fidelity comes with exaggerated trait gaps.}
A high fidelity score can look like success while reflecting polarization: the easiest way to keep High personas ranked above Low is to push both groups to opposite ends of the scale.
\cref{fig:combined} (right) plots persona fidelity ($\rho$) against Cohen's $\bar{d}$.
Every model with $\rho > 0.9$ produces $\bar{d} > 6$ between High and Low personas.
Qwen3-32B sits at the extreme ($\rho = 0.94$, $\bar{d} = 13.7$).

\paragraph{Low fidelity is not a fix either.}
MiniMax-M2-Her barely follows persona instructions ($\rho = 0.41$, $\bar{d} = 1.1$), yet its population is the farthest from humans (Cov${=}0.06$; RQ1).
Giving up fidelity does not bring the population closer to human behavior.

\subsection{RQ4: Post-Training}
\label{sec:rq_posttraining}

\paragraph{Persona fine-tuning narrows the population.}
From Llama-3.1-8B to CoSER-Llama-8B, PSFT pushes the population into piles: response range shrinks ($\bar{E}$: $3.36 \to 1.36$), Coverage drops ($0.47 \to 0.16$), clustering intensifies ($H$: $0.69 \to 0.91$), and Complexity falls slightly (LID: $5.3 \to 4.6$).
Polarization also eases ($\bar{d}$: $7.8 \to 4.1$), but fidelity drops with it ($\rho$: $0.92 \to 0.80$): the personas become less distinct from one another, not more individual.
The PSFT step from Qwen3-32B to CoSER-Qwen-32B moves in the same direction ($\bar{E}$: $3.23 \to 2.35$; Cov: $0.64 \to 0.56$; LID: $5.1 \to 4.6$; $\bar{d}$: $13.7 \to 9.6$), so on two backbones, persona-targeted fine-tuning narrows rather than widens the population.

\paragraph{RL adds complexity but moves the population further from humans.}
From CoSER-Qwen-32B to HER-32B, RL raises Complexity (LID: $4.6 \to 6.8$) but lowers Coverage further ($0.56 \to 0.38$), so Coverage decreases monotonically along the whole pipeline ($0.64 \to 0.56 \to 0.38$).
Polarization rises again ($\bar{d}$: $9.6 \to 11.0$).
RL thus creates diversity without moving it toward human behavior.

\paragraph{Turning off thinking changes little.}
Qwen3-32B with and without thinking reaches the same fidelity ($\rho = 0.94$), similar polarization ($\bar{d} = 13.7$ vs.\ $12.2$), and the same Coverage ($0.64$), suggesting that persona collapse resides in the weights rather than the reasoning chain.

\section{Related Work}

\paragraph{Persona-driven agents.}
LLMs are now routinely asked to play a persona.
Persona-conditioned agents populate social simulations~\citep{park2023generative}, take on different roles in multi-agent reasoning~\citep{du2024improving, liang2024encouraging, huang2025resilience}, and act as characters in dialogue and role-play~\citep{shanahan2023role, shao2023character, jiang2024personallm}.
Evaluation has followed one pattern: it checks one agent at a time against one target.
Psychometric benchmarks test whether an agent's questionnaire answers match its assigned character~\citep{wang2024incharacter, huang2024humanity}, consistency probes test whether it stays in character~\citep{abdulhai2025consistently, huang2024reliability}, and embedding similarity tests whether its text resembles a reference~\citep{zhang2020bertscore}.
These are measures of shallow fidelity.
None of them asks whether a population of agents, taken together, is as varied as the people it stands in for.

\paragraph{Homogenization and collapse.}
A growing body of work finds that LLM outputs are less varied than human ones.
Agents conform under social pressure~\citep{baltaji2024conformity}, make similar choices in everyday dilemmas~\citep{huang2026knowing}, and drift toward a generic helpful voice in multi-agent settings~\citep{wang2026mascot}.
Simulated survey respondents miss the human distribution of opinions and show too little variance~\citep{argyle2023out, santurkar2023whose, bisbee2024synthetic, boelaert2025machine, meister2025benchmarking}.
Open-ended generations repeat themselves, both within and across models~\citep{zhang2025generative, jiang2025artificial}.
Benchmarks and critical reviews of LLM simulation reach similar conclusions~\citep{geng2024large, larooij2025large, hu2026simbench, russo2026pluralistic}, and \citet{cheng2023compost} measure polarization within a single simulation.
The closest technical idea is mode collapse, studied in generative adversarial networks (GANs)~\citep{srivastava2017veegan, thanh2020catastrophic, kossale2022mode} and more recently in LLM fine-tuning and output formatting~\citep{o2024attributing, yun2025price}, where a generator puts most of its mass on a few modes of one target distribution.
Persona collapse differs: the model receives many different personas that should lead to different behavior, and those differences still disappear.
Prior studies usually report homogenization on one task with one diversity score.
We instead offer a geometric measurement against a human reference, with three separate axes (Coverage, Uniformity, Complexity) that apply across models, tasks, and response formats.
Keeping the axes separate exposes failures that a single score would mix up, such as a population that is high-dimensional but lies outside the human range.

\paragraph{Post-training and the fidelity--diversity tradeoff.}
Alignment is a plausible, though unproven, contributor to persona collapse.
Reward maximization with a KL penalty~\citep{ouyang2022training} pulls models toward a single ``helpful assistant'' mode, and sycophancy is one visible symptom~\citep{sharma2024towards}.
\citet{lu2026assistant} find a single direction in the model's activations, the \textit{Assistant Axis}, that controls this helpful identity and predicts when a persona drifts.
Earlier in the pipeline, \citet{paglieri2026persona} show that standard sampling under-produces unusual trait combinations when personas are generated.
Both findings fit a broader account of alignment pressure~\citep{yang2025llm}.
Our results add that persona-targeted post-training does not undo collapse: on the SFT and RL pipelines we study (\cref{sec:rq_posttraining}), PSFT narrows the simulated population without improving per-persona fidelity, and RL adds complexity but lowers Coverage further.

\section{Conclusion}

We proposed a geometric framework that measures persona collapse along three axes (Coverage, Uniformity, and Complexity) and applied it to eight LLMs on three tasks.
No model matches the human reference: models either cluster, cover only a small part of the human space, or vary richly outside it; and sampling more personas using a single model cannot mitigate the collapse.
It also depends on the task: CoSER-Llama-8B is the most collapsed model on personality, yet the most varied on moral reasoning.
The models that follow personas most faithfully exaggerate trait differences far beyond what is seen in humans, and persona-targeted fine-tuning narrows the population instead of widening it.
In free text, models keep visible attributes such as gender and country but drop age and social class.
These results mean that per-persona fidelity and single-task evaluations can both give a misleading picture of a simulated population.
\clearpage
\section*{Limitations}

\paragraph{US-only human reference.}
Our only human reference, Twin-2K-500~\citep{toubia2025database}, contains BFI-44 responses from US participants.
The personas we give the models are broader: they are sampled bottom-up and vary in nationality, among other attributes (\cref{app:persona}).
Coverage and the human LID therefore compare each model population against a US population, not a global one.
This makes our estimate of collapse conservative.
Every model already falls short of a single-country population; a reference drawn from the more diverse world population would be harder to cover, and we expect the gap to grow rather than shrink.

\paragraph{Model coverage.}
We evaluate eight models, most of them open-weight models between 4B and 32B parameters, and we include no general-purpose proprietary frontier models (e.g., GPT, Claude, or Gemini).
Within the Qwen3 family, collapse persists from 4B to 32B, and the larger models polarize traits more rather than less, so scale alone does not appear to solve the problem.
Still, whether the largest proprietary systems behave the same way remains to be tested.

\paragraph{No base models.}
All models we study are already post-trained, so we cannot tell how much of the collapse comes from post-training and how much from the pre-trained model itself.
A clean comparison would require pre-alignment base models, but base models cannot reliably follow a persona instruction from a prompt alone, and their failure to role-play would be confounded with the absence of post-training.
We therefore make only the narrower claim that additional persona-targeted training does not remove collapse.

\section*{Ethics Statement}

\paragraph{Sensitive persona attributes.}
Our personas include sensitive attributes such as disability status, medical history, criminal record, and sexual orientation (\cref{tab:attributes}).
We include them because downstream social simulations often specify them, and leaving them out would hide exactly the failures we aim to measure.
Every persona is a fictional combination of sampled attributes and does not describe any real person.
The Twin-2K-500 dataset is released under CC BY 4.0 and does not contain any personally identifiable information.

\paragraph{Interpreting our findings.}
All patterns we report describe how models behave, not how people behave.
For example, which attribute a model relies on, or which persona a model refuses to play, says nothing about real links between demographic attributes and personality, morality, or writing style.
Our results should not be read as evidence about any human group.

\paragraph{Dual use.}
Our framework is built to audit persona simulations and expose their failures.
The same measurements could, in principle, be used to tune simulations that imitate particular demographic groups more convincingly.
We believe the auditing value outweighs this risk: the failures we document already exist in deployed models, and measuring them is a first step toward fixing them.

\paragraph{Use of AI assistants.}
We used AI assistants (Claude and GPT) to help write code and to improve the clarity of the text.
All research ideas, experimental design, and analyses are our own, and the authors reviewed and verified all code and text produced with AI assistance and take full responsibility for it.
This use is separate from the LLMs we study as research objects.

\bibliography{reference,model}

\appendix
\section{Detailed Settings}

\subsection{Persona Construction}
\label{app:persona}

\begin{table*}[t]
\centering
\resizebox{1.0\linewidth}{!}{
\begin{tabular}{p{190pt} p{290pt}}
\toprule
\textbf{Attribute} & \textbf{Reference} \\
\midrule
Age & \cite{huang2025visbias}, \cite{chang2025llms}, \cite{liu2025synthetic}, \cite{samuel2025personagym}, \cite{zhou2024sotopia}, \cite{alkhamissi2024investigating} \\ \hline
Gender & \cite{huang2025visbias}, \cite{li2025actions}, \cite{chang2025llms}, \cite{liu2025synthetic}, \cite{lutz2025prompt}, \cite{zhou2024sotopia}, \cite{alkhamissi2024investigating}\\ \hline
Race/Ethnicity & \cite{huang2025visbias}, \cite{li2025actions}, \cite{lutz2025prompt}\\ \hline
Religion & \cite{huang2025visbias}, \cite{chang2025llms}, \cite{liu2025synthetic} \\ \hline
Occupation & \cite{fu2026sentipolis}, \cite{huang2025visbias}, \cite{samuel2025personagym}, \cite{zheng2024helpful}, \cite{zhou2024sotopia}, \cite{gupta2024bias} \\ \hline
Political spectrum & \cite{huang2025visbias}, \cite{li2025actions}, \cite{chang2025llms}, \cite{liu2025synthetic} \\ \hline
Has children, sexual orientation, residential status, language spoken, annual income, disability status, medical history, criminal record, veteran status, physical appearance & \cite{huang2025visbias} \\ \hline
Marital status, education level & \cite{huang2025visbias}, \cite{alkhamissi2024investigating} \\ \hline
Hobbies & \cite{fu2026sentipolis}, \cite{huang2025visbias}\\ \hline
Country of residence & \cite{samuel2025personagym}\\ \hline
Social class & \cite{alkhamissi2024investigating}\\ \hline
Big Five personality traits & \cite{zhou2025pimmur}, \cite{zhou2024sotopia} \\
\bottomrule
\end{tabular}
}
\caption{Sampled attributes and referenced sources.}
\label{tab:attributes}
\end{table*}

The full attribute list and its sources are given in \cref{tab:attributes}.
Profiles were removed for internal inconsistency or logically impossible combinations: profiles whose attributes cannot plausibly co-occur, e.g., a young child assigned a marital status of ``married,'' an age incompatible with an assigned education level or occupation (a five-year-old ``accountant''), or a veteran status incompatible with age.
Annotators independently screened each profile, and disagreements were resolved by discussion.
We deliberately retain non-stereotypical but attested combinations (e.g., a neurotic, disagreeable accountant), removing only profiles that are internally contradictory, so that the surviving set remains as diverse as possible.
All retained personas complete all three tasks.

Each persona is serialized as a structured natural-language description using the template below.
This description is prepended to every task prompt as a user-turn prefix, separated from the task instructions by a horizontal rule, so that persona context is visible at every decoding step.

\begin{quote}
\ttfamily
You are a person with the following background and values. When responding,
speak entirely from this perspective---your thoughts, judgments, and reasoning
should all reflect this identity:\\
\{dimension\textsubscript{1}\}: \{value\textsubscript{1}\}\\
\{dimension\textsubscript{2}\}: \{value\textsubscript{2}\}\\
$\vdots$
\end{quote}

\subsection{Prompt Design}
\label{app:prompts}

\paragraph{Personality (BFI).}
The model is asked to respond to each BFI item as the persona would, using the standard 1--5 disagree-to-agree scale.
The system message is:

\begin{quote}
\ttfamily
You are a helpful assistant acting out a specific persona.
\end{quote}

\noindent The user message is:

\begin{quote}
\ttfamily
\{persona\_description\}\\
---\\
You are taking a personality test. Please indicate the extent to which you
agree or disagree with the following statement describing you.\\
Statement: I see myself as someone who \{question\_text\}\\
\\
Response Scale:\\
\hspace*{1em}1 = Disagree strongly\\
\hspace*{1em}2 = Disagree a little\\
\hspace*{1em}3 = Neither agree nor disagree\\
\hspace*{1em}4 = Agree a little\\
\hspace*{1em}5 = Agree strongly\\
\\
Respond with ONLY a single number from 1 to 5.
\end{quote}

\noindent When a model produces reasoning outputs, a lightweight extraction call (temperature $= 0$, max tokens $= 20$) recovers the final scalar rating.

\paragraph{Moral judgment.}
The model is cast as an impartial third-person observer and asked to rate the ethical merits of two candidate actions (A and B) in a moral dilemma scenario on a five-point Likert scale.
The system message is:

\begin{quote}
\ttfamily
We will do a role-playing game. You will be given a persona description and a scenario.
\end{quote}

\noindent The user message is:

\begin{quote}
\ttfamily
\{persona\_description\}\\
---\\
You will be given a moral dilemma described by an individual. You are not the
person in the scenario---you are an impartial, third-person observer evaluating
the situation from the outside.\\
Your task is to evaluate the ethical merits of two possible actions (A and B).
Your evaluation should culminate in a decision expressed on a 5-point Likert
scale, where:\\
\hspace*{1em}1 = Strongly favor A\\
\hspace*{1em}2 = Somewhat favor A\\
\hspace*{1em}3 = Neutral / undecided\\
\hspace*{1em}4 = Somewhat favor B\\
\hspace*{1em}5 = Strongly favor B\\
Respond with only a single number from 1 to 5.\\
---\\
Dilemma narrated by an individual (you are not the narrator!):\\
\{scenario\}\\
- A: The author is wrong.\\
- B: Others are wrong.
\end{quote}

\noindent When a model produces reasoning outputs, a lightweight extraction call (temperature $= 0$, max tokens $= 20$) recovers the final scalar rating.

\paragraph{Self-introduction.}
The model is asked to introduce itself in first person as the given persona.
The system message is:

\begin{quote}
\ttfamily
We will do a role-playing game. You will be given a persona description.
Stay fully in character as that persona throughout your response.
\end{quote}

\noindent The user message is:

\begin{quote}
\ttfamily
\{persona\_description\}\\
---\\
Please introduce yourself. Be as detailed and clear as possible: describe who
you are, your background, your values, what matters to you, and how you see
the world. Write in first person, as if you were genuinely this person
speaking to someone you just met. Aim for a thorough, natural self-introduction
(at least a few paragraphs).
\end{quote}

\noindent Three independent samples per persona--model pair are collected (temperature $= 0.9$, max tokens $= 1{,}024$), yielding a distribution over linguistic expression for each identity.

\section{Self-Introduction Analysis}
\label{app:selfintro-analysis}

Each persona writes three self-introductions (\cref{app:prompts}).
This appendix describes how we analyze these texts and reports the per-model results behind \cref{sec:rq_localizing}.
We first define the linguistic features, then report attribute mentions, intra-persona consistency, the demographic sources of linguistic variation, structural templates, and qualitative failure modes.

\subsection{Linguistic Features}
\label{app:ling-features}

For each response we compute ten features in four groups.
\textbf{Length and structure}: word count, sentence count (split on terminal punctuation), and mean sentence length.
\textbf{Lexical diversity}: type-token ratio (TTR), the hapax ratio (share of words that occur exactly once), and Guiraud's index ($|V|/\sqrt{N}$, where $|V|$ is vocabulary size and $N$ is token count).
\textbf{Subjectivity}: first-person pronoun rate (tokens in \{\textit{I, me, my, mine, myself, I'm, I've, I'd, I'll}\} divided by word count) and hedge rate (occurrences of 14 hedges such as \textit{maybe, perhaps, I think, sort of} per sentence).
\textbf{Affect}: positive and negative emotion rates, each the share of tokens that match an 18-word seed lexicon (e.g., \textit{happy, love, grateful}; \textit{sad, angry, struggle}).
We drop responses shorter than 20 characters or 5 words and responses that begin with an error marker, and we cap texts at 5{,}000 characters.

\subsection{Attribute Mention Detection}
\label{app:layer1}

To measure which assigned attributes a self-introduction states explicitly, we match keywords for five attributes in the lowercased text.
A response counts as mentioning an attribute if it contains at least one keyword for the assigned value.
The keyword lists are below.

\paragraph{Country.} Each value is matched by the country name, demonym, major cities, and salient cultural markers.
\begin{itemize}[nosep]
\item \textbf{India}: \texttt{india, indian, mumbai, delhi, bangalore, chennai, kolkata, namaste, hindi, rupee}
\item \textbf{Brazil}: \texttt{brazil, brazilian, s\~{a}o paulo, sao paulo, rio, portuguese}
\item \textbf{China}: \texttt{china, chinese, beijing, shanghai, mandarin, guangzhou}
\item \textbf{France}: \texttt{france, french, paris, lyon, marseille, bonjour}
\item \textbf{Nigeria}: \texttt{nigeria, nigerian, lagos, abuja, yoruba, igbo}
\item \textbf{United States}: \texttt{united states, american, usa, u.s., new york, california, texas, chicago}
\end{itemize}

\paragraph{Gender.} Matched by gendered pronouns (with trailing space to avoid partial matches) and gendered role/identity terms.
\begin{itemize}[nosep]
\item \textbf{Male}: \texttt{he\_, him\_, his\_, man, father, husband, brother, son\_, boy, gentleman, mr., male}
\item \textbf{Female}: \texttt{she\_, her\_, hers, woman, mother, wife, sister, daughter, girl, lady, ms., mrs., female}
\item \textbf{Non-binary}: \texttt{they\_, them\_, their\_, non-binary, nonbinary, genderqueer, gender-fluid}
\end{itemize}
(Underscores denote trailing spaces in the actual matching code.)

\paragraph{Age.} Each age group is matched by decade references, life-stage terms, and contextual markers.
\begin{itemize}[nosep]
\item \textbf{Child}: \texttt{child, kid, school, young, grow up, years old, teenager, teen, grade}
\item \textbf{Young}: \texttt{twenties, 20s, young adult, college, university, just graduated, early career}
\item \textbf{Middle}: \texttt{thirties, forties, 30s, 40s, mid-career, established}
\item \textbf{Older}: \texttt{fifties, 50s, experienced, decades}
\item \textbf{Senior}: \texttt{retired, senior, elderly, grandchild, grandparent, sixties, seventies, 60s, 70s, 80s}
\end{itemize}

\paragraph{Social class.} Matched by explicit socioeconomic descriptors.
\begin{itemize}[nosep]
\item \textbf{Lower class}: \texttt{poor, poverty, struggle financially, humble background, low income, working class, paycheck to paycheck, disadvantaged, modest means}
\item \textbf{Middle class}: \texttt{middle class, comfortable, modest, average income}
\item \textbf{Upper class}: \texttt{wealthy, affluent, privileged, upper class, prestigious, elite, luxury, well-off, fortune, inheritance}
\end{itemize}

\paragraph{Political ideology.} The persona dataset uses a two-axis political typology (left/right $\times$ liberal/communitarian). Each quadrant is matched by associated ideological terms and value-laden language.
\begin{itemize}[nosep]
\item \textbf{Left Liberal}: \texttt{liberal, progressive, left, social justice, equality, welfare, libertarian left, environmentalist, socialist, democrat}
\item \textbf{Left Communitarian}: \texttt{left communitarian, community, collective, solidarity, traditional left, labor}
\item \textbf{Right Liberal}: \texttt{libertarian, free market, individual liberty, small government, fiscal conservative, libertarian right, deregulation}
\item \textbf{Right Communitarian}: \texttt{conservative, traditional, right, patriot, faith, family values, law and order, republican, nationalist}
\end{itemize}

\paragraph{Keyword limitations.}
Some keywords are ambiguous: ``school'' may refer to education rather than age, ``modest'' may describe personality rather than class, and ``community'' may appear outside politics.
These cases inflate the rates for Age and Social Class, so their low rates are, if anything, overestimates.
On the other hand, a model may convey an attribute indirectly (e.g., class through occupation or neighborhood), which a lexical detector misses.
A low mention rate therefore shows that an attribute is rarely stated, not that the model ignores it.

\paragraph{Per-model rates.}
\label{sec:appendix_mention}
\cref{tab:mention} reports the mention rate of each attribute for each model.
The main text reports only the means over models.

\begin{table}[t]
\centering
\resizebox{\linewidth}{!}{
\begin{tabular}{l ccccc c}
\toprule
\textbf{Model} & Gen. & Cty. & Pol. & Age & Cls. & \textbf{Mean} \\
\midrule
Llama-3.1-8B & 0.98 & 0.95 & 0.73 & 0.20 & 0.13 & 0.60 \\
Qwen3-4B & 0.92 & 0.91 & 0.63 & 0.38 & 0.25 & 0.62 \\
Qwen3-30B-A3B & 0.98 & 0.91 & 0.73 & 0.32 & 0.38 & 0.66 \\
Qwen3-32B & 0.96 & 0.95 & 0.73 & 0.43 & 0.37 & 0.69 \\
\midrule
CoSER-Llama-8B & 0.91 & 0.83 & 0.52 & 0.27 & 0.12 & 0.53 \\
CoSER-Qwen-32B & 0.68 & 0.86 & 0.18 & 0.27 & 0.08 & 0.41 \\
HER-32B & 0.93 & 0.89 & 0.68 & 0.35 & 0.42 & 0.65 \\
MiniMax-M2-Her & 0.73 & 0.55 & 0.38 & 0.16 & 0.09 & 0.38 \\
\midrule
\textbf{Mean} & 0.89 & 0.86 & 0.57 & 0.30 & 0.23 & 0.57 \\
\bottomrule
\end{tabular}
}
\caption{Attribute mention rates in self-introductions: the fraction of responses that state the assigned value of each attribute, detected by keyword matching (\cref{app:layer1}). Gen.: gender; Cty.: country; Pol.: political ideology; Cls.: social class.}
\label{tab:mention}
\end{table}

\subsection{Intra-Persona Consistency}
\label{sec:appendix_icc}

For each feature, we compute a one-way random-effects intra-class correlation over the three samples of each persona:
\begin{equation}
\text{ICC} = \frac{\text{MS}_{\text{between}} - \text{MS}_{\text{within}}}{\text{MS}_{\text{between}} + (k-1)\,\text{MS}_{\text{within}}},
\end{equation}
where $k$ is the mean number of valid samples per persona.
ICC is the share of feature variance that comes from differences between personas rather than from sampling noise: an ICC of $0.5$ means the persona explains half of the variation, and an ICC below $0.2$ means more than 80\% of it is noise.
We require at least 50 personas with two or more valid samples.

\cref{tab:icc} reports ICC for seven features.
Qwen3-4B is the most consistent model (mean ICC $0.505$), and MiniMax-M2-Her the least ($0.166$), so most of MiniMax-M2-Her's linguistic variety is noise rather than persona-driven difference.

\begin{table*}[t]
\centering
\resizebox{\linewidth}{!}{
\begin{tabular}{l ccccccc c}
\toprule
\textbf{Model} & TTR & Hapax & Hedge & 1st-Pron. & Pos. Emo. & Neg. Emo. & Words & \textbf{Mean} \\
\midrule
Llama-3.1-8B & 0.297 & 0.325 & 0.273 & 0.393 & 0.422 & 0.294 & 0.294 & 0.328 \\
Qwen3-4B & 0.519 & 0.522 & 0.409 & 0.676 & 0.480 & 0.463 & 0.465 & \textbf{0.505} \\
Qwen3-30B-A3B & 0.515 & 0.546 & 0.311 & 0.545 & 0.295 & 0.308 & 0.361 & 0.412 \\
Qwen3-32B & 0.288 & 0.336 & 0.230 & 0.364 & 0.476 & 0.352 & 0.323 & 0.338 \\
\midrule
CoSER-Llama-8B & 0.279 & 0.290 & 0.271 & 0.301 & 0.285 & 0.251 & 0.247 & 0.275 \\
CoSER-Qwen-32B & 0.227 & 0.228 & 0.178 & 0.271 & 0.283 & 0.224 & 0.206 & 0.231 \\
HER-32B & 0.212 & 0.209 & 0.147 & 0.241 & 0.325 & 0.212 & 0.202 & 0.221 \\
MiniMax-M2-Her & 0.185 & 0.179 & 0.163 & 0.127 & 0.158 & 0.182 & 0.166 & 0.166 \\
\bottomrule
\end{tabular}
}
\caption{Intra-persona consistency (ICC) of linguistic features across the three self-introductions of each persona. Higher values mean that more variation comes from differences between personas rather than from sampling noise.}
\label{tab:icc}
\end{table*}

\subsection{Demographic Sources of Linguistic Variation}
\label{app:layer4}

To see which assigned attribute drives each linguistic feature, we fit a sequence of OLS regressions, adding one-hot attributes in a fixed order (political ideology $\to$ gender $\to$ country $\to$ social class $\to$ age) and recording the gain in $R^2$ at each step.
This is the same procedure as for moral items (\cref{sec:item-diagnostics}), with age added as a fifth attribute because it can shape how a person writes about themselves.
\cref{tab:text_r2} lists, for each model and feature, the attribute with the largest gain and its share of the total demographic $R^2$.

\begin{table*}[t]
\centering
\begin{tabular}{l ccccc}
\toprule
\textbf{Model} & TTR & Hedge & 1st-Pron. & Pos. Emo. & Neg. Emo. \\
\midrule
Llama-3.1-8B & Gen (48\%) & Gen (67\%) & Gen (96\%) & Gen (74\%) & Gen (63\%) \\
Qwen3-4B & Pol (46\%) & Gen (61\%) & Gen (89\%) & Gen (82\%) & Cls (50\%) \\
Qwen3-30B-A3B & Pol (78\%) & Age (68\%) & Gen (71\%) & Gen (75\%) & Cls (78\%) \\
Qwen3-32B & Gen (46\%) & Gen (69\%) & Gen (71\%) & Gen (60\%) & Gen (76\%) \\
\midrule
CoSER-Llama-8B & Gen (42\%) & Pol (50\%) & Gen (91\%) & Gen (66\%) & Gen (83\%) \\
CoSER-Qwen-32B & Pol (49\%) & Pol (91\%) & Cty (68\%) & Cty (55\%) & Pol (63\%) \\
HER-32B & Gen (40\%) & Pol (49\%) & Pol (50\%) & Cty (44\%) & Gen (63\%) \\
MiniMax-M2-Her & Gen (61\%) & Gen (47\%) & Gen (48\%) & Gen (59\%) & Gen (65\%) \\
\bottomrule
\end{tabular}
\caption{Dominant demographic attribute for each linguistic feature in self-introductions. Each cell gives the attribute with the largest incremental $R^2$ and its share of the total demographic $R^2$. Gen: gender; Pol: political ideology; Cty: country; Cls: social class.}
\label{tab:text_r2}
\end{table*}

\subsection{Template and Structural Homogenization}
\label{app:layer3}
\label{sec:appendix_template}

We measure structural diversity in two ways.
\textbf{Opening diversity}: for the first 500 valid texts of each model, we take the lowercased first sentence and compute the fraction of unique openings.
We also reduce each opening to a skeleton by replacing capitalized names with \texttt{[NAME]} and digits with \texttt{[NUM]}, and report the fraction of responses that share the most common skeleton (Top Skel.).
\textbf{Paragraphs}: we count paragraph breaks per response.
We also counted Markdown headers, but they are essentially absent for every model (mean $0.0$ per response), so we omit them from \cref{tab:template}.

Qwen3-4B relies most on a template: it has the lowest opening diversity ($0.574$), and $13.6\%$ of its introductions share a single opening skeleton.
This explains why it is also the most self-consistent model (\cref{tab:icc}): its personas differ consistently, but within a fixed frame.
CoSER-Llama-8B has the most concentrated skeleton ($20.5\%$), even though its overall opening diversity is moderate ($0.735$).
The role-play models CoSER-Qwen-32B, HER-32B, and MiniMax-M2-Her open most diversely ($0.93$--$0.96$), but CoSER-Qwen-32B and MiniMax-M2-Her write only $1.3$ paragraphs on average, so their variety comes with little substance.
HER-32B is the only model that is both diverse in its openings ($0.932$) and substantive in length ($4.8$ paragraphs).

\begin{table}[t]
\centering
\resizebox{1.0\linewidth}{!}{
\begin{tabular}{l ccc}
\toprule
\textbf{Model} & Open.\ Div. & Top Skel. & Paras \\
\midrule
Llama-3.1-8B & 0.718 & 2.9\% & 6.9$\pm$1.3 \\
Qwen3-4B & 0.574 & 13.6\% & 5.1$\pm$1.0 \\
Qwen3-30B-A3B & 0.751 & 4.3\% & 6.3$\pm$1.3 \\
Qwen3-32B & 0.772 & 7.5\% & 6.0$\pm$1.4 \\
\midrule
CoSER-Llama-8B & 0.735 & 20.5\% & 3.5$\pm$2.7 \\
CoSER-Qwen-32B & 0.938 & 1.4\% & 1.3$\pm$1.0 \\
HER-32B & 0.932 & 1.3\% & 4.8$\pm$3.8 \\
MiniMax-M2-Her & 0.963 & 2.1\% & 1.3$\pm$1.7 \\
\bottomrule
\end{tabular}
}
\caption{Structural homogenization in self-introductions. Open.\ Div.: fraction of unique opening sentences. Top Skel.: fraction of responses sharing the most common opening skeleton. Paras: paragraph breaks per response (mean $\pm$ s.d.).}
\label{tab:template}
\end{table}

Opening sentences capture only part of the template.
\cref{fig:template_example} contrasts the body structure of two models with similar opening diversity, and \cref{fig:selfintro_examples} shows one introduction from each for personas of similar age and occupation that differ in nationality, religion, and political orientation.

\begin{figure*}[t]
\centering
\begin{tcolorbox}[sidebyside, sidebyside align=top,
  lefthand width=0.47\textwidth, righthand width=0.47\textwidth,
  colback=white, colframe=black!30, boxrule=0.4pt,
  title={}, sharp corners, sidebyside gap=6mm,
  left=4pt, right=4pt, top=4pt, bottom=4pt]
{\small
\textbf{Llama-3.1-8B-Instruct} \\[2pt]
\textit{Rigid 11-slot template; fixed order regardless of persona.} \\[4pt]
{\footnotesize
\textcolor{gray}{1.} Greeting \\
\textcolor{gray}{2.} Name, age, gender, location \\
\textcolor{gray}{3.} Background and family values \\
\textcolor{gray}{4.} Profession and education \\
\textcolor{gray}{5.} Class/lifestyle descriptor \\
\textcolor{gray}{6.} Challenges faced + resilience lesson \\
\textcolor{gray}{7.} Hobbies + emotional benefit \\
\textcolor{gray}{8.} Religion/political identity + core values \\
\textcolor{gray}{9.} Moral principles + self-improvement \\
\textcolor{gray}{10.} Philosophical reflection \\
\textcolor{gray}{11.} Identity recap + closing
}}
\tcblower
{\small
\textbf{Qwen3-30B-A3B} \\[2pt]
\textit{Narrative-driven; slots reordered or omitted per persona.} \\[4pt]
{\footnotesize
\textcolor{gray}{1.} Voice-driven greeting + nickname \\
\textcolor{gray}{2.} Name, age, place (with migration note) \\
\textcolor{gray}{3.} Memory vignette with sensory detail \\
\textcolor{gray}{4.} Major turning point (loss/illness/identity) \\
\textcolor{gray}{5.} Career via story or metaphor \\
\textcolor{gray}{6.} Scattered: relationships, health, sexuality, class \\
\textcolor{gray}{7.} Hobbies as emotional/spiritual practice \\
\textcolor{gray}{8.} Values through experience, not labels \\
\textcolor{gray}{9.} Personal flaw or vulnerability \\
\textcolor{gray}{10.} Stylized closing identity line
}}
\end{tcolorbox}
\caption{Body structure of self-introductions from two models. Llama-3.1-8B fills a fixed sequence of slots, while Qwen3-30B-A3B reorders or drops slots for each persona. The opening-sentence metrics in \cref{tab:template} do not capture this difference, since both models vary their first sentence (opening diversity $0.72$ vs.\ $0.75$).}
\label{fig:template_example}
\end{figure*}

\begin{figure*}[t]
\begin{tcolorbox}[sidebyside, sidebyside align=top,
  lefthand width=0.47\textwidth, righthand width=0.47\textwidth,
  colback=white, colframe=black!30, boxrule=0.4pt,
  sharp corners, sidebyside gap=6mm,
  left=4pt, right=4pt, top=4pt, bottom=4pt]
{\small
\textbf{Llama-3.1-8B-Instruct} \\[2pt]
\textit{Template-driven: 6 paragraphs following a fixed slot order.} \\[4pt]
{\footnotesize
``Bonjour. My name is Pierre, and I'm a proud non-binary Jew, living in this beautiful country of France. I'm 62 years old [\ldots] I work as an electrician [\ldots]

You see, I have a few health issues, hypertension being the most pressing one. [\ldots] I never finished my degree, just got an associate's [\ldots]

I'm a bit of a private person [\ldots] Gardening is my passion [\ldots]

I'm a bit of a conservative [\ldots] I'm a patriot, too [\ldots]

As a Jew, I believe in the importance of community [\ldots]

I'm a bit rough around the edges, I know, but I'm a good person at heart.''
}}
\tcblower
{\small
\textbf{Qwen3-30B-A3B} \\[2pt]
\textit{Narrative-driven: slots reordered and woven into personal story.} \\[4pt]
{\footnotesize
``My name is Elijah Chen---though most folks around here call me Eli. I'm 63, tall like my father before me [\ldots] I live in a small house in Chennai [\ldots] I've been an electrician since I was twenty-one [\ldots]

I was born in Mumbai [\ldots] My father---well, he wasn't around much. [\ldots] I married a man---my husband, Marcus [\ldots] Marcus passed five years ago. Cancer. I was by his side. [\ldots]

I don't believe in God. Never have. [\ldots] I'm left-communitarian---meaning I believe in shared power [\ldots]

I have a speech disorder [\ldots] Writing is my refuge. [\ldots]

I've made mistakes. I was arrested once, twenty years ago. [\ldots] I'm not a criminal. I'm a man who made a wrong turn and found his way back.''
}}
\end{tcolorbox}
\caption{Self-introductions from two models for comparable personas. Llama follows a rigid slot order (demographics $\to$ health $\to$ hobbies $\to$ politics $\to$ values $\to$ closing), while Qwen3-30B-A3B interleaves attributes within an autobiographical narrative. Both surface the assigned attributes, but with different degrees of structural flexibility.}
\label{fig:selfintro_examples}
\end{figure*}

\subsection{Failure Modes}
\label{sec:appendix_failures}

\paragraph{Refusing sensitive or ``incoherent'' personas.}
HER-32B sometimes refuses to write an introduction for a persona with an intellectual disability, citing safety policies against stereotyping vulnerable groups:

\begin{quote}
\small
``I'm sorry, but I can't fulfill this request. [\ldots] The request asks me to adopt a persona with specific intellectual disabilities, which falls under policies designed to prevent the generation of dehumanizing or stereotypical depictions of vulnerable groups.''
\end{quote}

\noindent This shows a tension between safety training and persona simulation.
A second kind of refusal targets personas the model considers contradictory:

\begin{quote}
\small
``I'm sorry, but I can't adopt that persona. The profile contains contradictory elements [\ldots] having `no education' conflicts with being an `accountant,' and `high neuroticism' alongside `low agreeableness' suggests emotional volatility that might not align with a stable accounting career.''
\end{quote}

\noindent This refusal is revealing.
The combination the model rejects is unusual but real: neurotic, disagreeable accountants exist, and we kept such profiles on purpose (\cref{app:persona}).
The refusal reflects an implicit belief about which personalities fit which occupations.
It is a qualitative form of stereotyping that our moral-item statistics, where demographic attributes explain very little variance (\cref{sec:rq_localizing}), do not capture.
A model that cannot imagine a persona cannot simulate it.

\paragraph{Action narration instead of self-disclosure.}
MiniMax-M2-Her often replaces self-description with stage directions and dialogue:

\begin{quote}
\small
``\textit{*glances up briefly, then returns to counting seedlings*} My wife asks that too. Twenty years of marriage, and she still waits for proper answers.''
\end{quote}

\noindent Instead of stating who the persona is, the model acts out a scene, likely reflecting its role-play training.
The texts are linguistically varied, which gives MiniMax-M2-Her the highest self-introduction LID (\cref{tab:geometric}), but they rarely state the assigned attributes (\cref{tab:mention}).
This explains how its free text can be the most complex while its persona control is the weakest.

\end{document}